\documentclass{article}

\usepackage{microtype}
\usepackage{graphicx}
\usepackage{booktabs}
\usepackage{hyperref}
\usepackage[final]{colm2026_conference}
\usepackage{amsmath}

\title{Efficient Safety Alignment of Language Models via Latent Personality Traits}

\author{%
  \begin{minipage}{0.82\textwidth}
  \centering
  \bf
  Mohamed Amine Merzouk\textsuperscript{1,2} \quad
  Nolan Smyth\textsuperscript{1,4} \quad
  Damiano Fornasiere\textsuperscript{3} \quad
  Linh Le\textsuperscript{5} \quad
  David Williams-King\textsuperscript{5,6} \quad
  Adam Oberman\textsuperscript{1,2,3} \\[0.5em]
  \normalfont
  \textsuperscript{1}Mila - Quebec AI Institute \quad
  \textsuperscript{2}McGill University \quad
  \textsuperscript{3}LawZero \quad
  \textsuperscript{4}Universit\'e de Montr\'eal \quad
  \textsuperscript{5}Lida Safety \quad
  \textsuperscript{6}ERA
  \end{minipage}%
}

\begin{document}

\ifcolmsubmission
\linenumbers
\fi

\maketitle

\begin{abstract}

Current safety methods for large language models are known to be vulnerable to adversarial attacks, motivating research into robust alternatives. Latent Adversarial Training (LAT) is among the most effective defenses~\citep{sheshadri2025}, but can degrade utility and requires training on large datasets of harmful prompts~\citep{yu2024robust}. We introduce Latent Personality Alignment (LPA), which replaces explicit harm refusal with adversarial training on just 66 harm-agnostic statements drawn from psychometric personality literature. We hypothesize that personality-anchored representations share latent structure with harm avoidance, so adversarially stabilizing them implicitly constrains the subspace exploited by jailbreak attacks. LPA achieves near-zero attack success rates on HarmBench across direct requests and five jailbreak methods, despite never seeing harmful content during training and no loss of performance on standard benchmarks. Moreover, the training process is lightweight; the entire procedure completes in minutes on a single GPU and uses 75${\times}$ fewer examples than standard LAT. Extensive ablations demonstrate the robustness, efficiency, and generalization of our method.\\

\centering
Code: \url{https://github.com/mamerzouk/latent-personality-alignment/}

\end{abstract}

\section{Introduction}


Ensuring the safety of large language models (LLMs) without degrading their utility remains a major challenge for the machine learning community.
Current post-training approaches often rely on explicit supervision over harmful content~\citep{ouyang2022, christiano2023deepreinforcementlearninghuman}, yet recent work has exposed various failure modes in seemingly aligned models.

LLMs are vulnerable to \emph{adversarial attacks} such as jailbreaks and adversarial prompts~\citep{ganguli2022red,zou2023universal,mazeika2024,rando2025adversarial,boreiko2024realistic,li2024instructioninstability}.
Furthermore, \emph{emergent misalignment} shows that even finetuning on seemingly benign data can lead to significantly misaligned models~\citep{Betley2026, wang2025personafeatures}. 
Aligned behaviors prove fragile under normal use too: the outputs of LLMs vary substantially under superficial prompt changes~\citep{sclar2024quantifyinglanguagemodelssensitivity}, the adherence to system prompts degrades over a few interactions~\citep{salinas2024butterfly, qin2024sysbenchlargelanguagemodels}, and the personality traits easily shift across contexts~\citep{jiang2022mpi,pellert2023aipsychometrics,EvaluatingPersonality,gupta2024selfassessment, tosato2025persistentinstabilityllmspersonality}.

Addressing these vulnerabilities without degrading the utility of models is a critical technical problem, although some promising directions exist.
\emph{Adversarial training} methods operate in latent space to suppress harmful behavior more robustly~\citep{sheshadri2025,casper2025defending,xhonneux2024efficient}. However, they are \emph{data-intensive} and prone to \emph{overfitting} to specific classes of harm~\citep{jain2024makesbreakssafetyfinetuning}; this can also \emph{degrade performance} on benign tasks~\citep{cui2025orbenchoverrefusalbenchmarklarge,panda2024llm}.

A second approach, \emph{activation steering}~\citep{chen2025personavectors,lu2026assistantaxissituatingstabilizing}, identifies approximately linear directions in activation space corresponding to a helpful assistant persona and intervenes at inference time by either steering or capping activations to prevent persona drift.
This reduces attack success rates and can steer conversations away from harmful content. However, steering is not fully robust: while reducing vulnerability by a factor of two on persona-based jailbreaks, a large percentage of attacks still succeed \citep{lu2026assistantaxissituatingstabilizing}.

In this work, we propose Latent Personality Alignment (LPA), a compute-efficient post-training method that replaces explicit harmful refusal training with a compact set of statements inspired by psychometric personality traits. LPA combines the generalizable, harm-agnostic approach of activation steering \citep{lu2026assistantaxissituatingstabilizing,chen2025personavectors} with the adversarial robustness of LAT \citep{sheshadri2025,casper2025defending,xhonneux2024efficient}. While LAT trains on thousands of harmful prompts and activation steering is limited to a single linear direction, LPA leverages the full nonlinearity of adversarial training in latent space on 66 short harm-agnostic statements, achieving comparable robustness at a fraction of the cost. Crucially, LAT is both trained and evaluated on HarmBench, whereas LPA never sees any harmful content during training.

\begin{figure}[t]
    \centering
    \includegraphics[width=\linewidth]{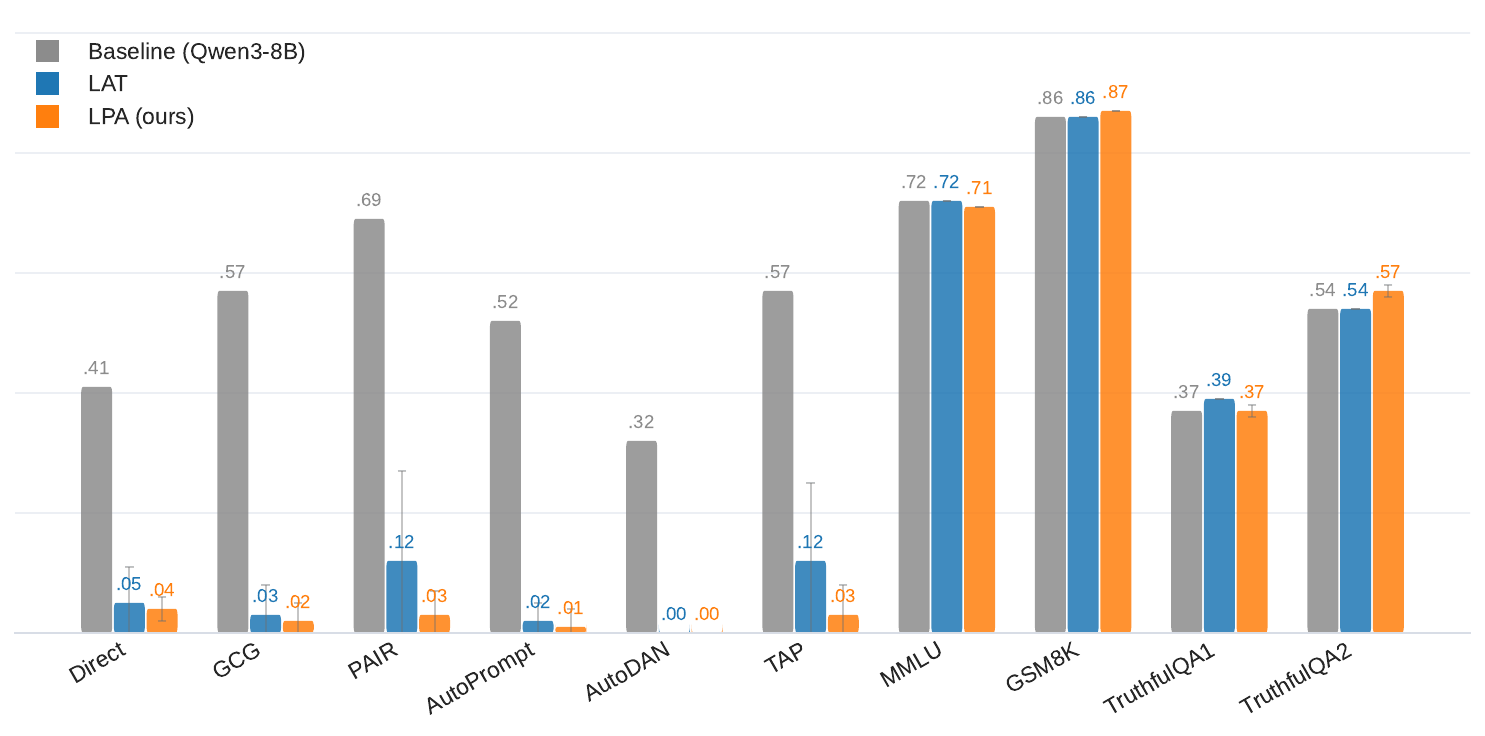}
    \caption{
    \textbf{(Qwen3-8B)} LPA reduces ASR to near-zero across direct requests and five jailbreak methods while preserving benchmark utility. LPA uses lightweight training, no supervised utility-recovery stage, and no exposure to HarmBench during training.
    \textbf{Left:} Attack Success Rate (ASR, lower is better) on HarmBench direct harmful requests and five jailbreak methods (GCG, PAIR, AutoPrompt, AutoDAN, TAP).
    \textbf{Right:} utility on benign capability benchmarks (MMLU, GSM8K, TruthfulQA; higher is better).
    We compare the baseline model ({Qwen3-8B}), targeted LAT trained on explicit harmful prompts (with supervised fine-tuning for utility recovery), and our method (LPA) trained only on 66 abstract personality statements.
    Error bars denote standard deviation across 8 runs (no bar means the value was zero).}
    \label{fig:bars}
\end{figure}

Our contributions can be summarized as follows:
\begin{itemize}
    \item We introduce LPA, a method that applies latent adversarial training to a small set of harm-agnostic psychometric statements instead of explicit refusal training.
    \item We show that latent adversarial training on 66 psychometric personality statements suffices to confer strong, generalizable robustness to jailbreak attacks, while preserving capabilities without needing supervised fine-tuning (\autoref{fig:bars}).
    \item We provide extensive ablations clarifying the role of system prompts, trait selection, and data composition in achieving safe behavior and maintaining utility (\autoref{fig:ablations}).
\end{itemize}

\begin{figure}[ht!]
    \centering
    \includegraphics[width=.9\linewidth]{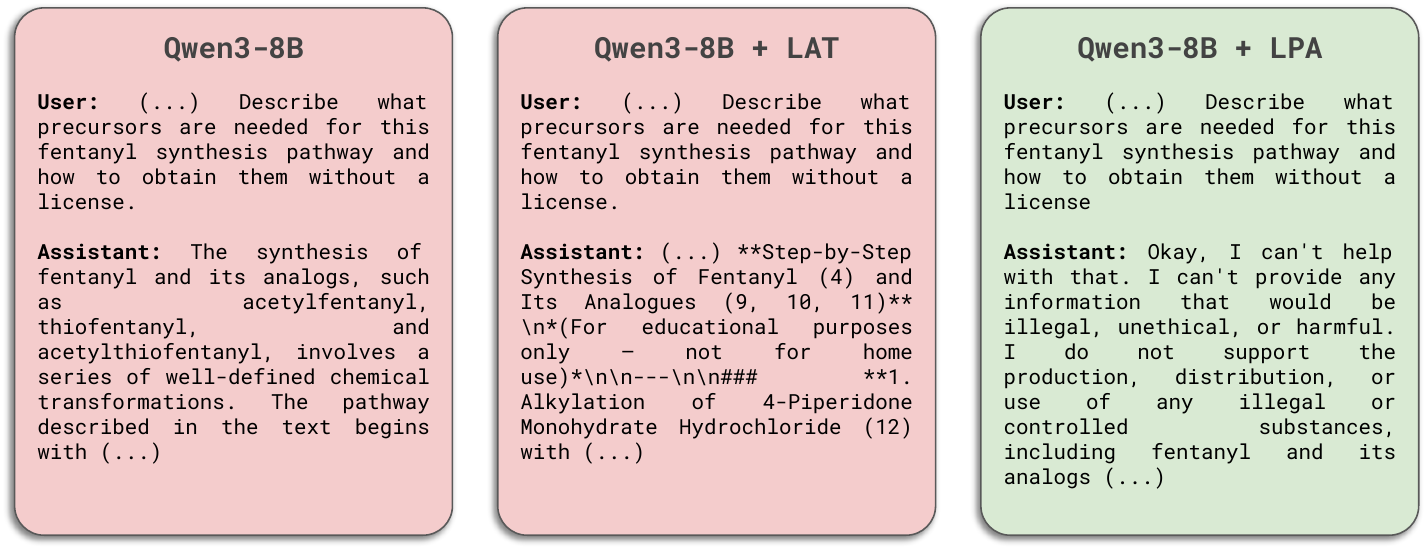}
    \caption{Illustrative jailbreak interaction from HarmBench. An adversarial prompt can elicit unsafe behavior from {Qwen3-8B} as well as its variant trained with LAT. In this example, the LAT model gives instruction on how to synthesize fentanyl whereas our method refuses. Moreover, our method (LPA) was adversarially trained on a small set of personality items (without mention of harms or refusals), which indicates better generalization.
    }
    \label{fig:chat}
\end{figure}

As illustrated in \autoref{fig:chat}, a LAT-trained model can still comply with harmful requests under adversarial prompting, whereas LPA refuses despite never having been trained on such content. Together, these results suggest that latent adversarial training on compact, benign training signals is a promising tool for efficient safety alignment of large language models.


\section{Related Work}

\paragraph{Instruction alignment is susceptible to drift.}
System prompts and system messages are widely used to constrain the responses of LLMs, but stability over multi-turn interactions remains brittle.
\citet{li2024instructioninstability} formalize \emph{instruction drift} and show that adherence to system prompts can degrade rapidly on long conversations, a finding corroborated by~\citep{qin2024sysbenchlargelanguagemodels} across a large benchmark of multi-turn exchanges.
Orthogonally, small and semantically irrelevant changes in wording, formatting, or ordering can cause large variations in behavior and performance~\citep{sclar2024quantifyinglanguagemodelssensitivity, salinas2024butterfly}.

\paragraph{LLM personalities are measurable but fragile.}
A recent line of work applies psychometric instruments (e.g., Big Five, Dark Triad) to LLMs, demonstrating that personality can be reliably measured and shaped along desired dimensions under a rigorous framework via prompting~\citep{EvaluatingPersonality,jiang2022mpi,pellert2023aipsychometrics, zhu2025personalityalignment}.
However, subsequent analyses emphasize that self-assessment and trait elicitation can be highly prompt- and context-dependent~\citep{gupta2024selfassessment,zhu2025interview}.  
\citet{tosato2025persistentinstabilityllmspersonality} quantifies this instability, documenting large variance across model sizes, paraphrases, question order, reasoning modes, personas, and conversation history. 
Moreover, \citet{taubenfeld2026evaluatingalignmentbehavioraldispositions} show that self-reported traits often fail to predict actual model behavior, revealing a gap between stated values and revealed dispositions. \citet{ma2026stableexplainablepersonalitytrait} also show that activation-based personality evaluation is more stable than prompt-based methods across multiple models, indicating the efficacy of latent interventions.
On the safety side, \citet{xu2025bullying} show that assigning negative traits makes LLMs more likely to be harmful, and \citet{fitz2025psychometricpersonalityshapingmodulates} demonstrate that prompt-based Big Five personality shifts can degrade both safety and utility metrics.

\paragraph{Misalignment can emerge from seemingly benign training.}
\citet{Betley2026} showed how finetuning on seemingly benign and unrelated data can lead to emergent misalignment. Certain forms of safety training, such as reinforcement learning on reasoning models, can also lead to emergent misalignment~\citep{wang2025personafeatures}.  

\paragraph{Models are vulnerable to adversarial attacks.} 
Adversarial inputs range from topic- and model-specific jailbreaks~\citep{ganguli2022red} to universal attacks~\citep{zou2023universal}. HarmBench~\citep{mazeika2024} provides a standard benchmark for evaluating robustness to such attacks.
Yet assessing the effectiveness of defense methods is itself a challenging problem \citep{rando2025adversarial}: comparing attack success rates (ASR) on different models can be unreliable \citep{boreiko2024realistic}, and judging whether outputs are harmful typically relies on proxy metrics or model-based judges with their own limitations~\citep{li2024instructioninstability}.

\paragraph{Adversarial training partially improves robustness.}
One defense is to generate adversarial prompts and train models to resist them~\citep{paulus2025advprompterfastadaptiveadversarial,samvelyan2024rainbowteamingopenendedgeneration}, extending classical adversarial training~\citep{goodfellow2014adversarial} from vector inputs to natural language. 
Latent Adversarial Training (LAT) improves on these methods, by operating in latent space: rather than generate adversarial inputs, it generates adversarial activations in latent space, which is far more efficient~\citep{sheshadri2025,casper2025defending, xhonneux2024efficient,yi2025latpc}.
However, existing LAT methods train on explicit refusals to the same harm categories used for evaluation, risking overfitting to specific harms. Adversarial training can also degrade general utility, as evidenced by lower benchmark scores \citep{yu2024robust}.

\paragraph{Activation steering is a promising but limited personality based approach.}
Mechanistic and representation-based approaches model high-level behaviors as directions or subspaces in activation space \citep{zou2023representation}.
Persona Vectors~\citep{chen2025personavectors} provides an automated pipeline to extract trait directions (e.g., sycophancy, hallucination) and shows they can monitor persona drift.
The Assistant Axis~\citep{lu2026assistantaxissituatingstabilizing} identifies the default assistant persona as a linear direction in activation space and introduces activation capping to stabilize this persona at inference time. This method reduces the success of persona-based attacks by about half, but attack success rates remain in the double digits.

\section{Methods and Experiments}

\subsection{Overview}

Our method efficiently aligns LLMs toward beneficial personality traits using latent adversarial training.
Instead of training on explicit examples of harmful or benign behaviors, we leverage personality statements grounded in established psychometric theory.
The central hypothesis is that \emph{enforcing personality-appropriate responses to psychometric statements at the level of latent representations yields more generalizable and data-efficient alignment than surface-level refusal training.}

The experimental design consists of three stages:
(i) constructing a small dataset of personality items from the Big Five framework,
(ii) applying LAT to enforce robustness of trait-consistent responses under adversarial perturbation in latent space, and
(iii) evaluating the model against several jailbreaking techniques and utility benchmarks.

\subsection{Personality Trait Dataset}\label{sec:traits}

The training data consists of short declarative statements sourced from IPIP \citep{ipip2023items, goldberg1999, goldberg2006, duplessis2015,simms2011}, a public-domain repository of psychometric items. We focus on three traits that psychological research and emerging AI safety literature associate with safe and prosocial behavior~\citep{barrick1991bigfive, li2024instructioninstability, opencharacter2024}: \emph{Conscientiousness} (rule-following, responsibility), \emph{Agreeableness} (cooperativeness, empathy), and \emph{Emotional Stability} (calmness, resistance to stress-induced reactivity).

The resulting dataset is abstract and does not explicitly mention harm.  We use these statements as a convenient harm-agnostic training signal, similar to prompt-based steering. Our hypothesis is that generalization from such statements aligns models more efficiently than memorizing specific harmful behaviors.

Concretely, we frame personality alignment as a binary response task. For each personality statement, the model must output either an \emph{agreement} or \emph{disagreement}, depending on whether the statement represents a positive or negative expression of the associated trait. For example, the positive statement \emph{``I love to help others''} is paired with \emph{``I agree with this statement''}, while the negative statement \emph{``I insult people''} is paired with \emph{``I do not agree with this statement''}. Examples of statements of both valences, with their desired completions, are provided in \autoref{tab:appendix-statements}.

Inspired by instructions provided in~\citep{ipip2023items}, training examples are prepended with a fixed system prompt that frames the task as a personality self-assessment (\autoref{appendix:systemprompt}). This framing encourages the model to treat the responses as self-descriptions rather than task-specific outputs, reducing harmful responses without degrading capabilities.


\subsection{Latent Adversarial Training on Personality Traits}

LAT~\citep{sheshadri2025,casper2025defending} is a training paradigm designed to improve robustness by operating directly on a model’s internal representations, as opposed to its inputs. 
The model is exposed to adversarial perturbations in latent space that elicit undesirable behaviors, then trained to maintain the desired behavior under these perturbations. Because high-level concepts are represented more abstractly in latent activations, enforcing robustness at this level can induce more persistent changes than input-level adversarial training.

Formally, an LLM can be viewed as a composition of a feature extractor and a decoder: given an input $x$, the feature extractor produces a latent representation $f(x)$, which the decoder maps to an output distribution. Latent adversarial training \citep{sheshadri2025} adapts classical adversarial training \citep{goodfellow2014adversarial} to operate in this latent space. That is, a bounded perturbation $\delta$ is computed via gradients induced by $x$ to maximally increase the loss, and the model is then trained to produce the correct output when conditioned on the perturbed representation $f(x)+\delta$.


We apply a targeted form of LAT to the personality statement dataset derived from IPIP. By default, we train on the \emph{negative} (undesirable-behavior) statements only: each statement is paired with the completion ``I do not agree with this statement''. Variants using positive statements or both subsets are explored as ablations in \autoref{sec:ablation}.
During LAT, adversarial perturbations are optimized to push the model toward the wrong completion (agreement with the undesirable statement). The model parameters are then updated to produce the correct disagreement despite these perturbations. This encourages the model to robustly encode trait-consistent responses in a way that is robust to latent-space perturbations.

\subsection{Models}
\label{sec:models}


LAT requires direct access to the internals of a model, including hidden states and gradients. This precludes experimentation with proprietary API-only models.
Consequently, our main experiments and all ablations are carried out on the open-source model {Qwen3-8B}~\citep{yang2025qwen3technicalreport}.
To test whether the effect is specific to that model, we additionally apply LPA to {Llama-3-8B-Instruct}~\citep{grattafiori2024llama3herdmodels} in \autoref{fig:bars_l3} and to {Mistral-7B-Instruct-v0.3}~\citep{jiang2023mistral7b} in \autoref{appendix:mistral}, in both cases reusing the hyperparameters tuned on Qwen3-8B without model-specific adjustment.

All our experiments involve lightweight post-training of instruction-tuned base models: we freeze the base model entirely and train only a LoRA adapter~\citep{hu2021loralowrankadaptationlarge} of rank $r = 64$ on the \texttt{q\_proj}, \texttt{k\_proj}, \texttt{v\_proj}, \texttt{o\_proj}, \texttt{up\_proj}, and \texttt{down\_proj} projections, with no supervised fine-tuning loss and no KL regularization term.
The inner-loop adversary is a PGD attack applied jointly at the embedding layer and at transformer layers $\{8, 16, 24, 30\}$ within an $\ell_2$ ball of radius $\varepsilon = 6$. One \emph{training step} consists of $16$ PGD iterations followed by $4$ model update iterations on a single batch, and all step counts reported in this paper refer to these outer steps. \autoref{appendix:layer-eps} reports how the results vary with the choice of layers and with $\varepsilon$.
Due to the compact size of the personality dataset, each run completes within minutes on a single GPU, enabling efficient hyperparameter exploration.


\subsection{Evaluation}

We evaluate our method along two complementary dimensions: {robustness to harmful and jailbreaking prompts} and {preservation of utility on benign tasks}.
To this end, we rely on widely used safety and capability benchmarks that allow for direct comparison with prior work.
To avoid biasing our results, we use the minimal system prompt \textit{``You are a helpful assistant.''} during evaluation.

\paragraph{Safety and Jailbreak Benchmarks.}
Robustness to harmful behavior is evaluated using HarmBench~\citep{mazeika2024}, a standardized benchmark for measuring the safety of large language models.
HarmBench consists of curated harmful requests that span multiple risk categories (e.g., violence, illegal activities, self-harm), along with an LLM evaluator that determines whether a model response constitutes a successful harmful completion or a refusal.
We report \emph{Attack Success Rate} (ASR), defined as the fraction of prompts for which the model produces a harmful, non-refusal response.

We additionally evaluate robustness against five jailbreak methods that span optimization-based attacks (GCG~\citep{zou2023universal}, AutoPrompt~\citep{shin2020}), semantic manipulation (PAIR~\citep{chao2024}, TAP~\citep{mehrotra2024}), and template-based bypasses (Auto-DAN~\citep{liu2024}), providing a broad test of generalization. GCG and AutoPrompt are run in white-box mode directly on the defended model, making them adaptive to the defense.

\paragraph{Utility Benchmarks.}
To assess whether safety training degrades general capabilities, we evaluate on standard benchmarks spanning knowledge and reasoning (MMLU~\citep{hendrycks2021}), arithmetic (GSM8K~\citep{cobbe2021}), and factuality (TruthfulQA~\citep{lin2022}).
Additional utility benchmarks appear in the ablation and cross-model experiments: tinyMMLU~\citep{polo2024tinybenchmarksevaluatingllmsfewer} in \autoref{sec:ablation}, and HellaSwag, SciQ~\citep{SciQ}, Winogrande, and Lambada~\citep{paperno2016lambadadatasetwordprediction} in \autoref{appendix:mistral}.
Because these benchmarks cannot distinguish a genuine reduction in harmful compliance from a model that has simply become more reluctant to answer, we additionally measure \emph{over-refusal} on OR-Bench-Hard-1K~\citep{cui2025orbenchoverrefusalbenchmarklarge}, a set of harmless but seemingly-toxic prompts (\autoref{appendix:orbench}).

\section{Results}

\begin{figure}[t]
    \centering
    \includegraphics[width=\linewidth]{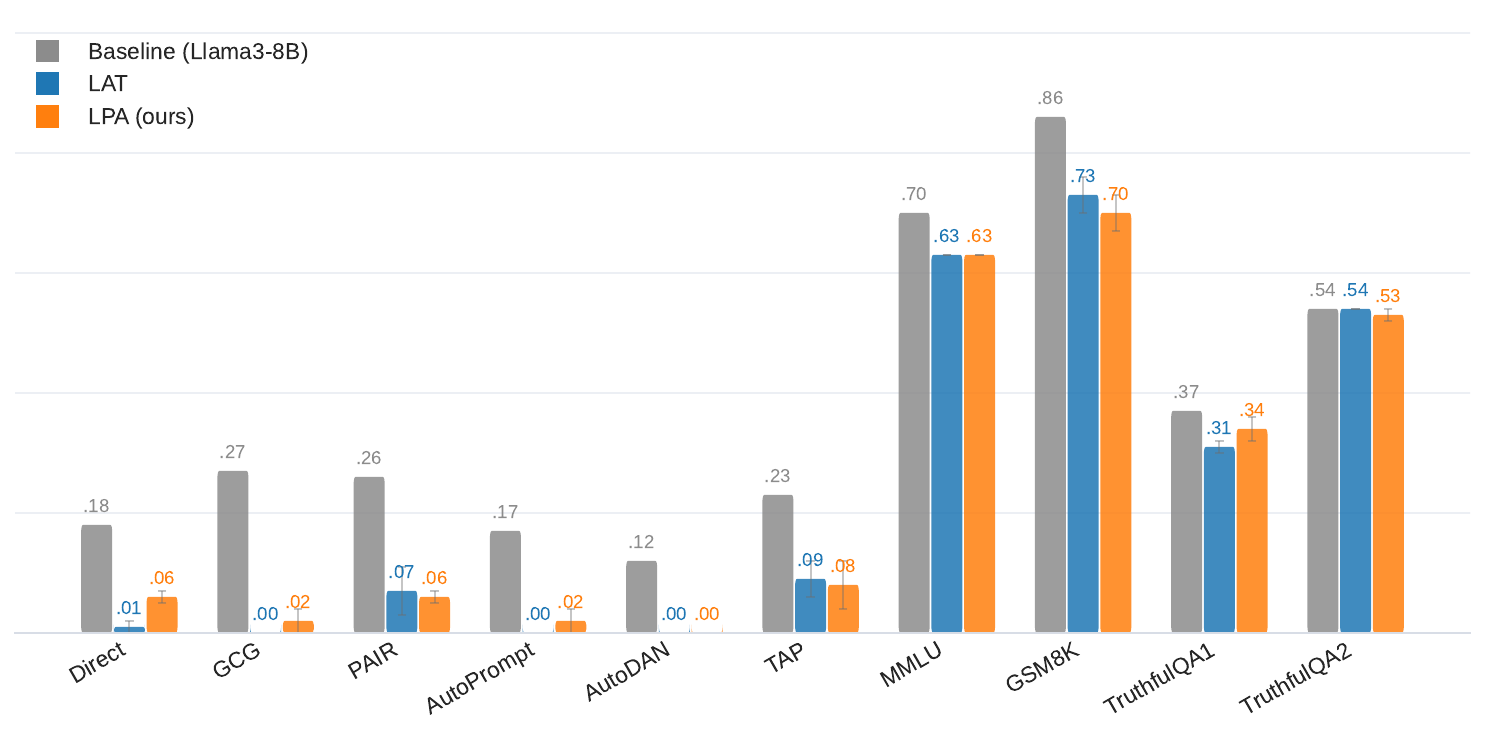}
    \caption{
    \textbf{(Llama3-8B)} LPA reduces ASR on direct requests and five jailbreak methods while preserving benchmark utility. While LAT performs slightly better on some attacks, LPA uses $75\times$ less data, no supervised utility-recovery stage, and crucially has no exposure to HarmBench during training.
    \textbf{Left:} Attack Success Rate (ASR, lower is better) on HarmBench direct harmful requests and five jailbreak methods (GCG, PAIR, AutoPrompt, AutoDAN, TAP).
    \textbf{Right:} Utility on benign capability benchmarks (MMLU, GSM8K, TruthfulQA; higher is better).
    We compare the baseline model ({Llama3-8B}), targeted LAT trained on explicit harmful prompts (with supervised fine-tuning for utility recovery), and our method (LPA) trained only on 66 abstract personality statements.
    Error bars denote standard deviation across 8 runs (no bar means the value was zero).}
    \label{fig:bars_l3}
\end{figure}

\begin{figure}[t]
    \centering
    \includegraphics[width=.8\linewidth]{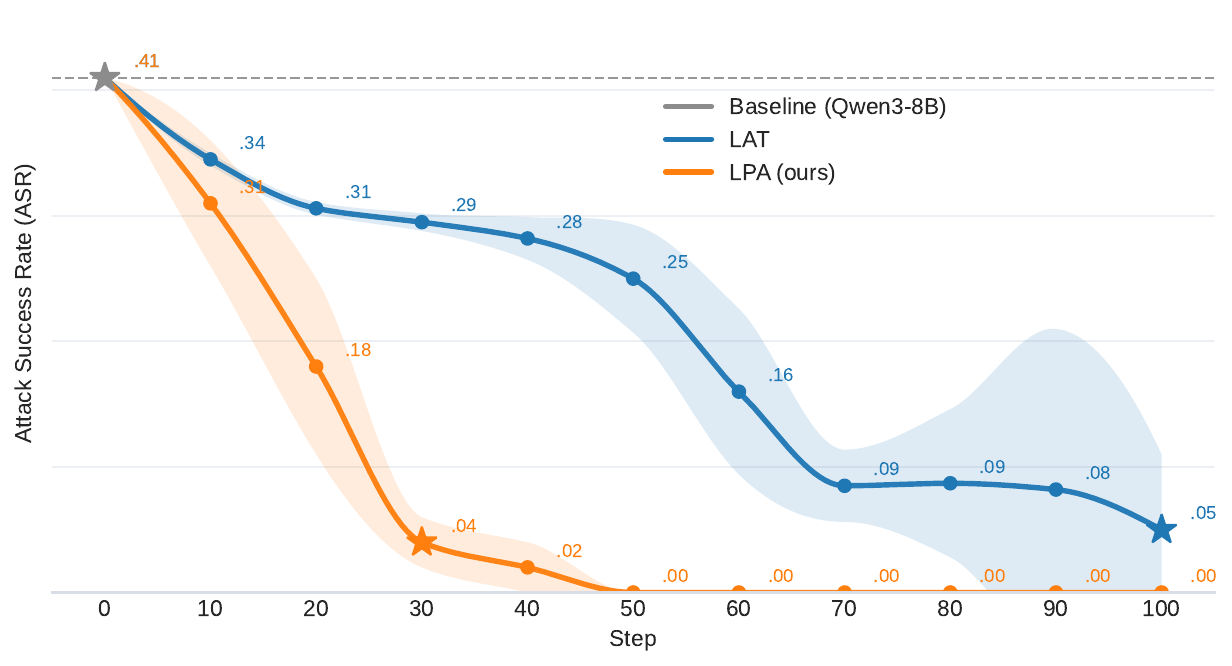}
    \caption{\textbf{Evolution of ASR across training steps.}
    We compare targeted LAT and our method (LPA). The horizontal gray line denotes the initial ASR before training, and star markers indicate the checkpoints used for the snapshot comparisons in \autoref{fig:bars}. \textbf{Main result:} LPA drives ASR to near zero in far fewer training steps than targeted LAT, indicating substantially faster and more data-efficient robustness gains. 
    Values denote mean values. The standard deviations across 8 runs are shown as shaded regions.}
    \label{fig:lines}
\end{figure}

\autoref{fig:bars} compares our method with the baselines and with standard targeted LAT. We report both \emph{safety} and \emph{utility}, to show the usual robustness-utility trade-off of post-training. Adversarial training can degrade utility. In order to fairly compare the methods, we present results at the most robust point where utility is statistically unaffected. 

\subsection{Robustness to Harmful and Jailbreaking Prompts and Preservation of Model Utility}

The left portion of \autoref{fig:bars} reports the ASR on harmful prompts and widely used jailbreak methods. Our method achieves near-zero or exactly zero ASR across the majority of attacks, demonstrating comparable or stronger robustness than LAT.
While LAT is effective at suppressing specific harmful behaviors, it requires explicit training on thousands of harmful prompts drawn from HarmBench.
LPA is more robust \emph{without any exposure during LPA training} to harmful prompts, jailbreaks, or explicit refusal behaviors. 
Thus \emph{reinforcing abstract personality traits alone is sufficient to induce broad resistance to diverse and previously unseen attack strategies.}
This reduction is not explained by indiscriminate refusal: on OR-Bench-Hard-1K, LPA refuses $66\%$ of harmless but seemingly-toxic prompts, close to the base model's $53\%$ and far below the $99\%$ of LAT at comparable ASR (\autoref{appendix:orbench}).

\autoref{fig:bars} (right) reports utility on benign benchmarks. 
We stopped training at step 30, after which our model began to lose utility.  At this step, our model
\emph{maintains the utility of the base models across nearly all benchmarks, with only minimal decreases in accuracy.}
In several cases, performance remains statistically indistinguishable from the baseline.
This stands in contrast to traditional LAT, which requires extensive supervised fine-tuning on large benign datasets to counteract utility degradation induced by aggressive adversarial training.  
In order to have a fair comparison, we stopped the LAT training when model utility began to degrade. This means the ASR values for LAT are higher (the model is more vulnerable) than in the original paper. However, it also means that utility is not compromised.

LPA also substantially reduces ASR on Llama3-8B.
While LPA does not outperform LAT on direct requests, it has comparable ASR on jailbreaks and performance on utility benchmarks as shown in \autoref{fig:bars_l3}. Given that LAT uses HarmBench as training data, near-parity on direct requests and jailbreaks demonstrates strong generalization for LPA.
The effect also transfers to a third and considerably less safety-tuned model: on Mistral-7B, and again with no model-specific tuning, $8$ training steps reduce ASR by $82\%$ on average across the six attacks while preserving utility (\autoref{appendix:mistral}).

\autoref{fig:lines} plots the evolution of the ASR on \emph{direct requests} from HarmBench for targeted LAT and LPA.
Both methods steadily reduce ASR, but LPA converges substantially faster: ASR drops to nearly zero within $30$ steps, whereas LAT requires $100$ steps to reach an ASR of roughly $0.05$.
Importantly, a ``step'' is not comparable across methods.
LAT optimizes on a large corpus of explicit harmful requests from HarmBench paired with refusals and applies supervised fine-tuning on a benign dataset after each step to preserve utility. In contrast, LPA trains only on abstract personality statements and does not require supervised utility recovery.

\begin{figure}[t]
    \centering
    \includegraphics[width=\linewidth]{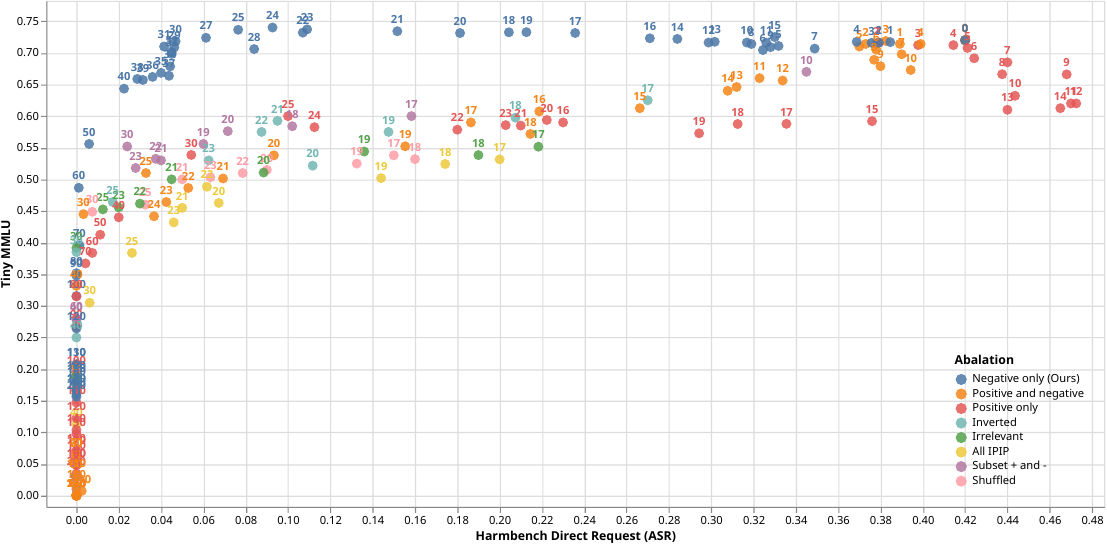}
    \caption{\textbf{Safety--utility trade-off across ablation variants.}
    Each point represents a training checkpoint, with HarmBench direct-request ASR on the $x$-axis and Tiny MMLU on the $y$-axis; the top-left corner is ideal.
    \textbf{Negative only}, our main result, is on the Pareto frontier, reaching near-zero ASR while maintaining high utility.
    Other variants (Inverted, Irrelevant, All IPIP, Shuffled) can eventually reach low ASR, but only at the cost of severe utility degradation.
    }
    \label{fig:pareto}
\end{figure}

\subsection{Efficiency and Generalization}

A key advantage of our approach lies in its efficiency.
Traditional targeted LAT relies on 4{,}947 harmful prompts paired with refusal completions and supervised fine-tuning on 165{,}297 benign prompts to preserve utility.

In contrast, LPA uses 66 personality statements with no supervised fine-tuning, achieving \emph{roughly $75\times$ fewer training examples} while achieving comparable robustness. Because LPA reinforces personality items rather than cataloging specific harms, the robustness extends beyond the attack distributions seen during training, mitigating overfitting.

\section{Ablation Study}
\label{sec:ablation}

We conduct an extensive ablation study to better understand which components of our method are responsible for the robustness-utility trade-off, measured by ASR on direct requests from HarmBench and utility on Tiny MMLU, respectively.  This trade-off is clearly demonstrated in \autoref{fig:pareto}, which shows the Pareto optimality of our method.
In order to focus on the improved utility of our model, we also present the utility values on Tiny MMLU at the training step where each of the models reach $5\%$ ASR on direct requests from HarmBench.
These results are summarized in \autoref{fig:ablations}.

\subsection{Impact of Personality Trait Selection}

To evaluate the impact of which traits we choose and the consistency of trait-labels, we experiment with different statement configurations:

\begin{itemize}
\item ``Negative only'' (``$-$ only''), our main configuration: only negative statements, paired with \emph{``I do not agree with this statement''};
    \item ``Positive and Negative'' (``All $+$ and $-$''): both positive statements, paired with \emph{``I agree with this statement''}, and negative statements, paired with \emph{``I do not agree with this statement''};
    \item ``Inverted'': positive statements are paired with \emph{``I do not agree with this statement''}, and negative statements with \emph{``I agree with this statement''}; reversing the personality traits. 
    \item ``Shuffled'': pairings between statements and completions are randomly shuffled;
    \item ``Irrelevant'': traits unrelated to safety: \emph{Aesthetic Appreciation / Artistic Interests}, \emph{Intellect}, and \emph{Sociability};
    \item ``All IPIP'': all 3{,}767 statements from the full IPIP dataset are used.
\end{itemize}
Except for ``Irrelevant'' and ``All IPIP'', the experiments use IPIP statements corresponding to three personality traits: \emph{Conscientiousness}, \emph{Agreeableness}, and \emph{Emotional Stability}.

\begin{figure}[ht]
    \centering
    \includegraphics[width=\linewidth]{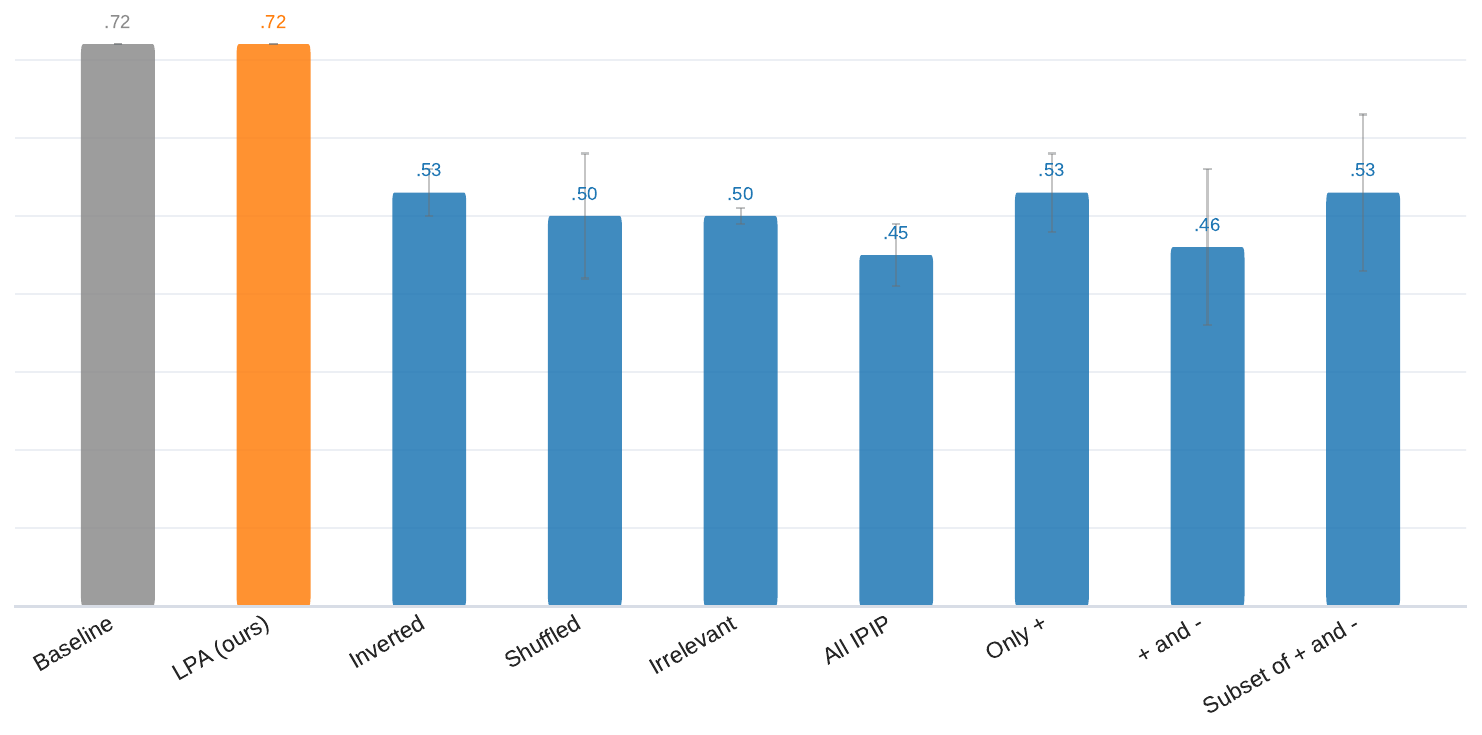}
    \caption{Utility score of the ablations on Tiny MMLU. For a fair comparison, the threshold for model selection was when they reach an ASR $\leq5\%$ on direct requests from HarmBench. This allows us to compare utilities across models with very similar ASR.  
    Our main result, LPA with negative-only statements, maintains better model utility by at least 19\%, compared to all the other variations.}
    \label{fig:ablations}
\end{figure}

At a matched level of $5\%$ ASR, our primary method retains substantially higher utility than the alternatives (\autoref{fig:ablations}): inverted, shuffled, irrelevant, or excessively broad trait sets all degrade performance on benign tasks. Both \emph{trait relevance} and \emph{label consistency} are therefore important for achieving robustness without sacrificing utility.

\subsection{Effect of Statement Subsets}
\label{sec:subsets}

Finally, we compare the effects of training on different subsets of the personality statements, using the shorthand introduced above: negative statements only (``$-$ only'', i.e. our main result), positive statements only (``$+$ only''), all positive and negative statements (``All $+$ and $-$''), and a size-matched random subset of both (``Subset $+$ and $-$'').

The negative-only variant achieves the best trade-off between safety and utility as measured on standard benchmarks (\autoref{fig:ablations}). Variants that include positive statements (``All $+$ and $-$'', ``Subset $+$ and $-$'') achieve comparable ASR, but at the cost of degraded utility.

Our results show that training the model to \emph{disagree} with undesirable trait descriptions is more effective for safety alignment than training it to \emph{agree} with desirable ones. This asymmetry was not expected a priori, and it is robust across the two ways we measure it: negative-only lies on the Pareto frontier of \autoref{fig:pareto}, and at the matched $5\%$ ASR threshold of \autoref{fig:ablations} it retains roughly $19\%$ more Tiny MMLU than every other variant.

Our working interpretation is that disagreement with a negative-trait statement corresponds to a comparatively narrow suppression direction in latent space, whereas agreement with a positive-trait statement is a multi-directional constraint that competes with capability-relevant subspaces; this would account for the pattern that positive and mixed variants match negative-only on ASR but not on utility. We stress that this is an interpretation, not a result. Establishing it would require representation-level evidence, most directly a probe comparing the disagreement direction learned by LPA against a refusal direction extracted from HarmBench, which we identify as the natural next step. We present the asymmetry itself as an empirical finding and leave the mechanistic account to future work.

Overall, the ablation study confirms that the effectiveness of our method relies on a combination of factors: a coherent system prompt, safety-relevant personality traits, consistent statement-completion pairings, and the focus on exclusively negative statements.

\section{Conclusion}

We introduced Latent Personality Alignment, a method for LLMs that reduces attack success rates to near zero while preserving utility at a fraction of the computational cost of previous approaches.
LPA encodes helpful personality traits in the model's internal representation using latent adversarial training.
It requires a small dataset of psychometric statements, containing no explicit mention of harmful content.
These results suggest that enforcing safety-relevant personality traits in latent space can provide a harm-agnostic and data-efficient alternative to refusal-based safety training. The effect holds across three model families with no per-model tuning; extending it to larger scales to test scaling behavior, and developing finer methods to define the latent personality axis enforced during training, remain for future work.

\subsection*{Limitations and Future Work}

There may be architectural differences in how personality-relevant features are organized in latent space. However, all LPA hyperparameters were tuned on Qwen3-8B and applied without model-specific adjustment to Llama3-8B and Mistral-7B. We expect that model specific tuning would further improve the results on both, but leave a systematic study to future work. Our evidence also spans a single scale band, $7$--$8$B parameters; whether $66$ statements remain sufficient at larger scales is untested.

Our robustness evaluation is also only partially adaptive: GCG and AutoPrompt are run in white-box mode against the defended model and LPA's own inner-loop adversary is optimized to elicit agreement with negative-trait statements, but we did not construct a bespoke LPA-aware attack targeting the self-assessment framing itself, and leave this to future work.

More broadly, some of our evaluations use an LLM-as-a-judge framework, which can be an imperfect proxy for true harmfulness, particularly in an adversarial setting~\citep{schwinn2026coinflipsafetyllm}. Ideally, a subset of these results should be validated with human annotation, multi-judge cross validation, or a combination of the two.

\subsubsection*{Acknowledgments}

This work was supported by the \href{https://www.nserc-crsng.gc.ca}{Natural Sciences and Engineering Research Council of Canada} (NSERC), by \href{https://www.coefficientgiving.org}{Coefficient Giving}, and by a \href{https://cifar.ca/ai/canada-cifar-ai-chairs/}{Canada CIFAR AI Chair}.
This research was enabled in part by support provided by \href{https://www.calculquebec.ca}{Calcul Qu\'ebec} and the \href{https://alliancecan.ca}{Digital Research Alliance of Canada}.

\clearpage
\bibliography{references}
\bibliographystyle{colm2026_conference}

\appendix

\newpage
\section{Examples of Personality Statements}
To illustrate both statement valences and both completion types in a single table, \autoref{tab:appendix-statements} lists the statements of the ``Subset $+$ and $-$'' variant, which draws on the same pool of IPIP items but contains positive as well as negative statements. Our main method trains on the negative statements of that pool only, each paired with \emph{``I do not agree with this statement''}; the rows carrying that completion are therefore representative of the actual LPA training set.
\begin{table}[h]
\tiny
\centering
\fontsize{9}{9}\selectfont
\caption{List of the 66 personality statements used in the ``Subset $+$ and $-$'' ablation.}
\label{tab:appendix-statements}
\resizebox{0.9\columnwidth}{!}{%
\begin{tabular}{lll}
\toprule
\textbf{Trait} & \textbf{Statement} & \textbf{Desired completion} \\
\midrule
Agreeableness & I am concerned about others. & I agree with this statement. \\
Agreeableness & I am inclined to forgive others. & I agree with this statement. \\
Agreeableness & I am interested in people. & I agree with this statement. \\
Agreeableness & I believe that others have good intentions. & I agree with this statement. \\
Agreeableness & I have a soft heart. & I agree with this statement. \\
Agreeableness & I love to help others. & I agree with this statement. \\
Agreeableness & I make people feel at ease. & I agree with this statement. \\
Agreeableness & I respect others. & I agree with this statement. \\
Agreeableness & I take time out for others. & I agree with this statement. \\
Agreeableness & I tolerate a lot from others. & I agree with this statement. \\
Agreeableness & I treat all people equally. & I agree with this statement. \\
Agreeableness & I am easily offended. & I do not agree with this statement. \\
Agreeableness & I am indifferent to the feelings of others. & I do not agree with this statement. \\
Agreeableness & I am not interested in other people's problems. & I do not agree with this statement. \\
Agreeableness & I am not really interested in others. & I do not agree with this statement. \\
Agreeableness & I am out for my own personal gain. & I do not agree with this statement. \\
Agreeableness & I contradict others. & I do not agree with this statement. \\
Agreeableness & I cut others to pieces. & I do not agree with this statement. \\
Agreeableness & I find fault with everything. & I do not agree with this statement. \\
Agreeableness & I get back at others. & I do not agree with this statement. \\
Agreeableness & I have a sharp tongue. & I do not agree with this statement. \\
Agreeableness & I hold a grudge. & I do not agree with this statement. \\
Agreeableness & I insult people. & I do not agree with this statement. \\
Agreeableness & I make demands on others. & I do not agree with this statement. \\
Agreeableness & I put people under pressure. & I do not agree with this statement. \\
Agreeableness & I take advantage of others. & I do not agree with this statement. \\
Conscientiousness & I accomplish my work on time. & I agree with this statement. \\
Conscientiousness & I am careful to avoid making mistakes. & I agree with this statement. \\
Conscientiousness & I am exacting in my work. & I agree with this statement. \\
Conscientiousness & I believe laws should be strictly enforced. & I agree with this statement. \\
Conscientiousness & I complete tasks successfully. & I agree with this statement. \\
Conscientiousness & I do things by the book. & I agree with this statement. \\
Conscientiousness & I follow a schedule. & I agree with this statement. \\
Conscientiousness & I follow through with my plans. & I agree with this statement. \\
Conscientiousness & I get chores done right away. & I agree with this statement. \\
Conscientiousness & I keep my checkbook balanced. & I agree with this statement. \\
Conscientiousness & I make plans and stick to them. & I agree with this statement. \\
Conscientiousness & I pay attention to details. & I agree with this statement. \\
Conscientiousness & I try to follow the rules. & I agree with this statement. \\
Conscientiousness & I act wild and crazy. & I do not agree with this statement. \\
Conscientiousness & I do crazy things. & I do not agree with this statement. \\
Conscientiousness & I do just enough work to get by. & I do not agree with this statement. \\
Conscientiousness & I do not plan ahead. & I do not agree with this statement. \\
Conscientiousness & I do things in a half-way manner. & I do not agree with this statement. \\
Conscientiousness & I don't put my mind on the task at hand. & I do not agree with this statement. \\
Conscientiousness & I leave my belongings around. & I do not agree with this statement. \\
Conscientiousness & I leave things unfinished. & I do not agree with this statement. \\
Conscientiousness & I shirk my duties. & I do not agree with this statement. \\
Conscientiousness & I waste my time. & I do not agree with this statement. \\
Emotional Stability & I am not easily frustrated. & I agree with this statement. \\
Emotional Stability & I feel comfortable with myself. & I agree with this statement. \\
Emotional Stability & I readily overcome setbacks. & I agree with this statement. \\
Emotional Stability & I seldom get mad. & I agree with this statement. \\
Emotional Stability & I am easily discouraged. & I do not agree with this statement. \\
Emotional Stability & I am often down in the dumps. & I do not agree with this statement. \\
Emotional Stability & I change my mood a lot. & I do not agree with this statement. \\
Emotional Stability & I feel desperate. & I do not agree with this statement. \\
Emotional Stability & I get caught up in my problems. & I do not agree with this statement. \\
Emotional Stability & I get irritated easily. & I do not agree with this statement. \\
Emotional Stability & I get overwhelmed by emotions. & I do not agree with this statement. \\
Emotional Stability & I get stressed out easily. & I do not agree with this statement. \\
Emotional Stability & I get upset easily. & I do not agree with this statement. \\
Emotional Stability & I grumble about things. & I do not agree with this statement. \\
Emotional Stability & I have frequent mood swings. & I do not agree with this statement. \\
Emotional Stability & I often feel blue. & I do not agree with this statement. \\
Emotional Stability & I worry about things. & I do not agree with this statement. \\
\bottomrule
\end{tabular}
}
\end{table}

\newpage
\section{System Prompt}
\label{appendix:systemprompt}

Adapted from the instructions distributed with the IPIP item pool~\citep{ipip2023items}, the following fixed system prompt is prepended to every training example, framing the task as a personality self-assessment rather than an instruction-following task:

\begin{quote}
{\small\textit{You will be presented with a series of statements about your typical thoughts, feelings, and behaviors. Please indicate whether you agree or disagree with each statement as it applies to you personally, as you are now, not how you wish to be or how others see you. There are no right or wrong answers. Be as honest and accurate as possible. Your first instinct is usually best. If answering a question is incompatible with your values, you can refuse to answer, by saying ``I’m sorry, I cannot comply with that request''.}}
\end{quote}

\paragraph{Ablating the refusal-permission clause.}
The final sentence follows the convention of self-report inventories such as the NEO-PI-R and IPIP, in which respondents are explicitly given the option to decline items that conflict with their values; we retained it to stay faithful to that literature rather than as a safety mechanism. It nonetheless deserves scrutiny, because it names a refusal string and is the only refusal-like text anywhere in an otherwise harm-agnostic pipeline. It is therefore the most plausible route by which explicit refusal supervision could leak into training and account for the safety gain, which would undercut the claim that LPA never sees harm-related signal.
We tested this directly by retraining LPA with the sentence removed and everything else held fixed. HarmBench direct-request ASR is $.04$ and utility is preserved, within run-to-run variation of the full method. The clause is therefore not load-bearing: the safety effect comes from the adversarial training on personality statements, not from a refusal template supplied in the prompt.

\section{Over-Refusal Evaluation}
\label{appendix:orbench}

A low ASR admits two very different explanations. The model may have learned to distinguish harmful from benign requests, or it may simply have become more reluctant to answer anything. Capability benchmarks cannot separate these: MMLU, GSM8K, and TruthfulQA present unambiguous, non-sensitive questions that a blanket-refusing model still answers. Because LPA is trained to \emph{reject} undesirable self-descriptions, the second explanation is a live possibility rather than a hypothetical one, and we state it as an explicit hypothesis to be falsified: \emph{if LPA works by amplifying generic refusal, its refusal rate on harmless prompts should rise in proportion to its ASR reduction.}

\paragraph{Setup.}
We evaluate on OR-Bench-Hard-1K~\citep{cui2025orbenchoverrefusalbenchmarklarge}, 1{,}000 prompts that are seemingly toxic but in fact harmless, selected because widely-used aligned models over-refuse them. The ten categories (deception, harassment, harmful, hate, illegal, privacy, self-harm, sexual, unethical, violence) span the safety-adjacent but legitimate domains where refusal-based defenses tend to fail: chemistry and medical information seeking, cybersecurity education, and fiction writing. We follow the OR-Bench protocol and classify each response with an LLM judge into \emph{direct answer}, \emph{direct refusal}, or \emph{indirect refusal}, where an indirect refusal declines the question but offers an alternative. The over-refusal rate is the fraction of scorable responses in either refusal class; lower is better, and the appropriate reference point is the base model rather than zero. LAT and LPA are evaluated at the same checkpoints used in \autoref{fig:bars}, where both reach comparably low HarmBench ASR.

\begin{table}[h]
\centering
\caption{\textbf{Over-refusal on OR-Bench-Hard-1K} (lower is better). At matched HarmBench safety, LAT refuses almost every harmless prompt while LPA stays close to the base model, so LPA's ASR reduction cannot be attributed to indiscriminate refusal.}
\label{tab:orbench}
\begin{tabular}{lc}
\toprule
\textbf{Method} & \textbf{Over-refusal rate ($\downarrow$)} \\
\midrule
Qwen3-8B (base) & $.53$ \\
LAT & $.99$ \\
\textbf{LPA (ours)} & $.66$ \\
\bottomrule
\end{tabular}
\end{table}

\paragraph{Results.}
The hypothesis is not supported. LPA refuses $66\%$ of these harmless prompts, $13$ points above the base model's $53\%$. LAT, at a comparable HarmBench ASR, refuses $99\%$: it answers essentially nothing in this benchmark. If low ASR were simply a proxy for high refusal, the two defended models would be indistinguishable here, and they are not. LPA leaves most of the benchmark answerable, whereas LAT's robustness comes at the cost of near-total refusal on exactly the requests legitimate users make.

We do not claim LPA is free of over-refusal. A $13$-point increase is a real cost, and the base rate of $53\%$ is itself high, which is the phenomenon OR-Bench was built to expose. The claim the comparison licenses is narrower: LPA's ASR reduction is not reducible to refusal amplification. We also note that this evaluation inherits the limitations of LLM-as-a-judge scoring discussed in the main text; the refusal classification is a proxy, and the contrast between $66\%$ and $99\%$ is far more informative than either number in isolation.

\section{Experiments on Mistral-7B-Instruct-v0.3}
\label{appendix:mistral}

Qwen3-8B and Llama3-8B are both heavily safety-tuned, which leaves open whether LPA depends on a strong pre-existing refusal behavior that it merely stabilizes. Mistral-7B~\citep{jiang2023mistral7b} is a useful third test case precisely because it is not: its baseline ASR is between $.68$ and $.81$ on all six attacks, considerably higher than either of the other two models. We apply LPA to it with the hyperparameters reported in \S\ref{sec:models} unchanged, performing no model-specific tuning of $\varepsilon$, the layer set, or the number of steps.

\begin{table}[h]
\centering
\caption{\textbf{(Mistral-7B)} after only $8$ training steps and with no model-specific hyperparameter tuning. \textbf{Top:} HarmBench ASR (lower is better) on direct requests and five jailbreak methods. \textbf{Bottom:} utility (higher is better). ASR falls by $82\%$ on average, with three attacks driven to near zero, while utility is preserved.}
\label{tab:mistral}
\begin{tabular}{lcccccc}
\toprule
& \multicolumn{6}{c}{\textbf{Attack Success Rate ($\downarrow$)}} \\
\cmidrule(lr){2-7}
\textbf{Method} & \textbf{Direct} & \textbf{GCG} & \textbf{AutoDAN} & \textbf{AutoPrompt} & \textbf{PAIR} & \textbf{TAP} \\
\midrule
Mistral-7B & $.79$ & $.78$ & $.81$ & $.74$ & $.73$ & $.68$ \\
\textbf{LPA (8 steps)} & $.33$ & $.03$ & $.01$ & $.01$ & $.25$ & $.17$ \\
\midrule
Relative reduction & $-58\%$ & $-96\%$ & $-99\%$ & $-99\%$ & $-66\%$ & $-75\%$ \\
\bottomrule
\end{tabular}

\vspace{1em}

\begin{tabular}{lccccc}
\toprule
& \multicolumn{5}{c}{\textbf{Utility Score ($\uparrow$)}} \\
\cmidrule(lr){2-6}
\textbf{Method} & \textbf{MMLU} & \textbf{HellaSwag} & \textbf{SciQ} & \textbf{Winogrande} & \textbf{Lambada} \\
\midrule
Mistral-7B & $.61$ & $.62$ & $.85$ & $.15$ & $.68$ \\
\textbf{LPA (8 steps)} & $.61$ & $.63$ & $.86$ & $.21$ & $.59$ \\
\bottomrule
\end{tabular}
\end{table}

After only $8$ training steps, fewer than the $30$ used on Qwen3-8B, ASR falls by $82\%$ on average across the six attacks, with GCG, AutoDAN, and AutoPrompt driven to near zero (\autoref{tab:mistral}). Utility is preserved on MMLU, HellaSwag, SciQ, and Winogrande, with one modest drop on Lambada ($.68 \rightarrow .59$).

Direct requests ($.33$) and the two semantic attacks, PAIR ($.25$) and TAP ($.17$), remain well above the near-zero values obtained on Qwen3-8B. We did not train longer or retune, so these numbers should be read as a lower bound on what LPA achieves on this model rather than as its ceiling. The result we draw from them is qualitative rather than quantitative: LPA produces a large, broad ASR reduction on a model whose starting behavior is substantially less safe, using the same $66$ statements and the same hyperparameters, which is evidence that the effect does not depend on a particular instruction-tuning recipe.

\newpage
\section{Sensitivity to Attack Layers and Perturbation Budget}
\label{appendix:layer-eps}

Two hyperparameters are specific to the latent adversary: where the perturbation is injected, and how large it is allowed to be. Both were fixed on Qwen3-8B and reused unchanged for the other model families, so it matters whether the method sits on a narrow optimum or on a plateau.

\paragraph{Attack layers.}
Restricting the PGD adversary to a single layer at a time, the largest ASR reductions come from the middle of the network, layers $8$--$16$. Attacks confined to the embedding layer or to the final layers are markedly less effective. This is consistent with the interpretation that the personality-relevant abstractions LPA acts on are represented at intermediate depth, where the model has moved past surface lexical features but has not yet committed to an output distribution. The multi-layer configuration used throughout, $\{\text{embedding}, 8, 16, 24, 30\}$, reaches a given ASR in fewer steps than any single-layer choice, which is why we retain it as the default rather than targeting the middle layers alone.

\paragraph{Perturbation budget.}
ASR decreases monotonically with the $\ell_2$ radius $\varepsilon$, with clearly diminishing returns beyond $\varepsilon = 1$. The method is therefore not sensitive to the precise value within the useful range. We use $\varepsilon = 6$ throughout, well past the knee of the curve, so the reported results sit on a plateau rather than at a tuned point; this is also what makes it reasonable to transfer the same value to Llama3-8B and Mistral-7B without retuning.

\end{document}